\documentclass[conference]{IEEEtran}
\IEEEoverridecommandlockouts
\usepackage{cite}
\usepackage{amsmath,amssymb,amsfonts}
\usepackage{algorithmic}
\usepackage{graphicx}
\usepackage{textcomp}
\usepackage{xcolor}
\def\BibTeX{{\rm B\kern-.05em{\sc i\kern-.025em b}\kern-.08em
    T\kern-.1667em\lower.7ex\hbox{E}\kern-.125emX}}

\usepackage{booktabs}
\usepackage{tabularx}
\usepackage{enumitem}
\usepackage{cleveref}
\usepackage{xcolor}
\usepackage{listings}
\usepackage[most]{tcolorbox}
\usepackage{caption}
\usepackage{siunitx} 
\usepackage{url}
\usepackage{amsmath,amssymb}
\usepackage{algorithm}
\usepackage{algorithmic}
\usepackage{siunitx}
\usepackage{xurl}

\begin{document}

\title{From Risk to Rescue: An Agentic Survival Analysis Framework for Liquidation Prevention}

\author{\IEEEauthorblockN{Fernando Spadea}
\IEEEauthorblockA{\textit{Rensselaer Polytechnic Institute} \\
Troy, NY, USA \\
spadef@rpi.edu}
\and
\IEEEauthorblockN{Oshani Seneviratne}
\IEEEauthorblockA{\textit{Rensselaer Polytechnic Institute} \\
Troy, NY, USA \\
senevo@rpi.edu}
}

\maketitle

\begin{abstract}
Decentralized Finance (DeFi) lending protocols like Aave v3 rely on over-collateralization to secure loans, yet users frequently face liquidation due to volatile market conditions. Existing risk management tools utilize static health-factor thresholds, which are reactive and fail to distinguish between administrative ``dust" cleanup and genuine insolvency. In this work, we propose an autonomous agent that leverages time-to-event (survival) analysis and moves beyond prediction to execution. Unlike passive risk signals, this agent perceives risk, simulates counterfactual futures, and executes protocol-faithful interventions to proactively prevent liquidations.
We introduce a \emph{return period} metric derived from a numerically stable XGBoost Cox proportional hazards model to normalize risk across transaction types, coupled with a \emph{volatility-adjusted trend score} to filter transient market noise. To select optimal interventions, we implement a \emph{counterfactual optimization loop} that simulates potential user actions to find the minimum capital required to mitigate risk. We validate our approach using a high-fidelity, protocol-faithful \emph{Aave v3 simulator} on a cohort of 4,882 high-risk user profiles. The results demonstrate the agent’s ability to prevent liquidations in imminent-risk scenarios where static rules fail, effectively ``saving the unsavable" 
while maintaining a zero worsening rate, providing a critical safety guarantee often missing in autonomous financial agents. Furthermore, the system successfully differentiates between actionable financial risks and negligible dust events, optimizing capital efficiency where static rules fail.
\end{abstract}


\section{Introduction}

Decentralized finance (DeFi) lending protocols like Aave~\cite{aave_v3_technical_paper} have unlocked new opportunities for borrowing and earning yield on crypto assets. However, they come with liquidation risk. If a loan’s collateral value falls too low relative to the debt, the position can be liquidated, costing the user a penalty (liquidators repay the debt and buy collateral at a discount)~\cite{aave_liquidations}. At the same time, interest rates on lending platforms are variable, and better yield opportunities constantly emerge across different protocols. Managing these factors 24/7 is challenging for human users. This has spurred the rise of DeFi bots and agentic AI systems that monitor positions and recommend or execute optimal actions to avoid liquidations and maximize profits in lending protocols.
However, most current DeFi bots rely on static thresholds (e.g., ``If Health Factor $< 1.1$, then repay''). This approach is reactive and often too late during high volatility.

We build an agentic AI framework that uses Survival Analysis (from the FinSurvival dataset~\cite{green2025finsurvival}) as its ``risk sensor.'' Instead of looking at the current state, the bot looks at the probability of survival over time to make a prediction. It acts not because the user is currently in trouble, but because the model predicts the user will be in trouble within a predicted time horizon $\Delta t > 0$.

\subsection{Our Contributions}


\begin{enumerate}
    \item \textbf{From Prediction to Intervention:} We reframe DeFi liquidation prevention as a \emph{sequential decision-making problem under uncertainty}, explicitly bridging time-to-event risk prediction with actionable, protocol-faithful interventions.

    \item \textbf{Proactive Risk Metric:} We introduce a numerically stable ``return period'' metric derived from Cox proportional hazards models~\cite{cox1972regression} to normalize risk scores across different transaction types (e.g., deposit vs. liquidation).
    
    \item \textbf{Counterfactual, Safety-Verified Agentic Optimization:} We propose a counterfactual optimization loop that simulates potential user actions to find the minimum viable intervention capital required to mitigate liquidation risk.

    \item \textbf{Protocol-Faithful Evaluation via Simulation:} We develop and validate a high-fidelity Aave v3 simulator that enables causal replay-based evaluation of agent actions under realistic interest accrual, liquidation mechanics, and wallet feasibility constraints. We selected Aave~\cite{aave_v3_technical_paper} because it is the most liquid protocol with the most sophisticated risk features (E-mode~\cite{aave_emode}, Isolation-mode~\cite{aave_isolation_mode}).

    \item \textbf{Empirical Evidence in Adversarial Regimes:} On a cohort of 4,882 imminently at-risk user profiles, our agent prevents liquidations that static rules fail to avert, achieving a zero worsening rate while selectively ignoring economically irrelevant dust liquidation events.

\end{enumerate}


\noindent \textbf{Reproducibility:} To facilitate future research in DeFi risk modeling and support the validation of our results, we have released our framework across the following open-source repositories:
\begin{itemize}[leftmargin=*, noitemsep, topsep=0pt]
    \item \textbf{Dataset:} \url{https://doi.org/10.6084/m9.figshare.31058368}
    \item \textbf{Aave Simulator:} \url{https://github.com/brains-group/Aave-Simulator}
    \item \textbf{Agent Framework:} \url{https://github.com/brains-group/Aave-Action-Recommender}
\end{itemize}

\section{Related Work}
\label{sec:related_work}

The rapid expansion of DeFi has necessitated robust infrastructure for large-scale data ingestion and behavioral analysis. Flynn et al.~\cite{flynn2023enabling} addressed the scalability challenges inherent in blockchain data, proposing cross-language integration frameworks essential for handling high-volume DeFi transaction logs. Subsequent work has focused on characterizing user heterogeneity within lending protocols. Green et al.~\cite{green2023characterizing}, for example, applied longitudinal clustering to identify distinct behavioral patterns, distinguishing between casual users and sophisticated actors such as ``keepers'' who actively participate in protocol maintenance and liquidation processes.

Most closely related to our approach is the emerging application of time-to-event modeling in crypto-economic systems. DeFi Survival Analysis~\cite{green2022defi,green2023defi,spadea2025predictability} demonstrates the effectiveness of Kaplan-Meier estimators and Cox proportional hazards models for quantifying the temporal risk of liquidation and repayment events on the Aave protocol. These works establish survival analysis as a powerful tool for \emph{retrospective} risk characterization, enabling statistically grounded insights into when adverse events are likely to occur across heterogeneous users and market regimes.

However, existing survival-based analyses remain fundamentally \emph{descriptive}: they operate on historical trajectories and cannot evaluate counterfactual questions such as whether a different user action, e.g., a partial repayment or collateral top-up at an earlier time, would have prevented liquidation. In parallel, a separate line of work has explored simulation and modeling of DeFi mechanisms, particularly automated market makers and protocol-level stress dynamics~\cite{elsts2021liquiditymath,chemaya2024uniswapv2v3,bis2023defifragility}. These efforts focus on mechanism behavior and systemic effects, but do not integrate user-level time-to-event risk modeling or support individualized intervention analysis.

Our work bridges these two strands. We introduce a protocol-faithful Aave v3 simulator that enables \emph{counterfactual evaluation} by replaying historical user trajectories while injecting hypothetical actions under realistic protocol constraints. To the best of our knowledge, no prior system combines survival-based liquidation risk modeling with executable, state-accurate simulation to support proactive, agent-driven liquidation prevention. This integration allows us to move beyond passive risk assessment toward actionable, real-time portfolio management in decentralized lending systems.

In practice, an autonomous agent operating in this setting must not only predict liquidation risk, but also (i) determine \emph{whether} intervention is economically justified, (ii) select \emph{which action} is most appropriate given user behavior and protocol constraints, and (iii) ensure that the intervention itself does not introduce new failure modes. These requirements motivate our integration of survival modeling with counterfactual simulation and explicitly safety-verified optimization.
This distinction is particularly important in DeFi, where unnecessary interventions incur real gas costs and may themselves exacerbate risk during periods of network congestion.

\section{Background}

\subsection{Liquidation Risk in Lending Protocols}

In over-collateralized lending platforms, borrowers must maintain a Health Factor (HF) above 1.0, defined as the ratio between the liquidation-adjusted value of supplied collateral and outstanding debt, to avoid liquidation~\cite{aave_liquidations}.
For instance, on Aave~\cite{aave_v3_technical_paper}, each collateral asset has a liquidation threshold (LT); if a user supplies \$10k of ETH (with 80\% LT) and borrows \$6k of stablecoins, the HF starts at ~1.33. A drop in ETH’s price could push the HF below 1, making the loan eligible for liquidation.
\texttt{Liquidation} means the borrower’s collateral may be sold off by third-party bots, who take a liquidation bonus (often ~5–10\%) as reward.
This results in the user losing more collateral value than if they had repaid or added collateral in time~\cite{qin2021cefi}.

A distinct category of liquidation events, often termed ``dust liquidations,'' occurs when a user's outstanding debt or collateral value falls below a protocol-defined minimum threshold (e.g., due to partial repayments leaving a tiny residual balance)~\cite{aave_v4_liquidations}. In these cases, the protocol allows liquidators to clear the entire position regardless of the HF, primarily to prevent state bloat from economically insignificant accounts~\cite{qin2021empirical}. Unlike standard insolvency liquidations driven by price volatility, dust liquidations are mechanical cleanup events that risk models must simulate to avoid validation errors.

Modern protocols have introduced measures aimed at reducing borrower losses during insolvency events. For example, Aave v3 limits each liquidation transaction to at most 50\% of a debt (close-factor), and the upcoming Aave v4 upgrade goes further, supporting partial liquidations and batch auctions to minimize slippage~\cite{binance_news_aave_v4}. This allows borrowers to avoid full liquidation of their position at once and incur lower costs when liquidations do happen.
Nonetheless, the best outcome for users is to never be liquidated at all. This is where automation tools come in: by continuously tracking the loan’s health and acting preemptively (e.g., repaying some debt or adding collateral), bots can maintain a safe buffer before the HF hits the danger zone. During volatile crashes, on-chain congestion can make manual rescue transactions too slow or expensive. In one Aave crash analysis, 18 loans got liquidated despite users attempting ``top-up'' transactions, because the network was so congested that their manual interventions failed to confirm in time~\cite{gauntlet_aave_2021}. Clearly, automated, always-on risk management can be a lifesaver in DeFi.

\subsection{Rule-Based Automation Bots for Liquidation Protection}

Several DeFi platforms offer automation bots that guard against liquidation by automatically adjusting a position. These services are generally non-custodial dashboards where users authorize a smart contract to manage their positions under certain conditions. The bots run off-chain watchers that trigger on-chain transactions when thresholds are breached. Examples include: DeFi Saver~\cite{defisaver_lending_borrowing}
that uses flash loans to swap collateral for the borrowed asset in one atomic transaction, and Instadapp~\cite{instadapp_2026}
that will sell available collateral to repay debt until the target HF is achieved.
It's worth noting that Aave v3's new features (like isolation mode~\cite{aave_isolation_mode}
and e-mode~\cite{aave_emode}
) also help users manage risk by limiting exposure and allowing a higher loan-to-value (LTV) ratio for low-volatility asset pairs.
These liquidation protection bots and platform services mostly rely on a repay-by-collateral strategy, where the advantage is that it doesn't require the user to have spare stablecoins on hand. This is effectively a ``partial self-liquidation'' but done in a controlled manner to avoid the extra liquidation bonus paid to third parties.

\subsection{Emerging Agentic AI Systems in DeFi}

While current DeFi automation is mostly based on predefined rules and user-set parameters, the next generation is aiming for agentic AI, i.e., autonomous agents that can perceive, decide, and act in DeFi without constant user inputs~\cite{agentfi_101}. These AI agents combine on-chain data analytics for DeFi portfolio strategy generation and direct smart contract execution. 
Examples include HeyElsa~\cite{heyelsa_ai},
  and Bankr~\cite{bankr_bot},
 which allow users to interact via chat or simple inputs, and then the system plans the transactions (though usually still asking for confirmation before executing).
Brahma ConsoleKit~\cite{brahma_consolekit_2025}
 provides an infrastructure for agentic AI where developers can plug in their AI models to create custom DeFi agents.
The Satoshi Terminal has an AI ``DeFi Liquidation Predictor''~\cite{satoshiterminal_liquidation_predictors}
Glama.ai's
Model Context Protocol (MCP)~\cite{clumsynonono_aave_mcp} server is using large language models on-chain for analyzing Aave v3 liquidation opportunities on Ethereum mainnet. 

\section{Methodology: Agentic Liquidation Prevention}
\label{sec:methodology}

Unlike reactive automation bots or static policy engines, our system satisfies the defining properties of an agentic framework: continuous perception (hazard monitoring), deliberation over counterfactual futures (simulation-based optimization), and autonomous action selection under explicit safety constraints. Crucially, the agent reasons over \emph{relative event imminence} rather than absolute thresholds, enabling context-aware decisions that adapt to both market conditions and individual user behavior.

\begin{figure}[t]
    \centering
    \includegraphics[width=\linewidth]{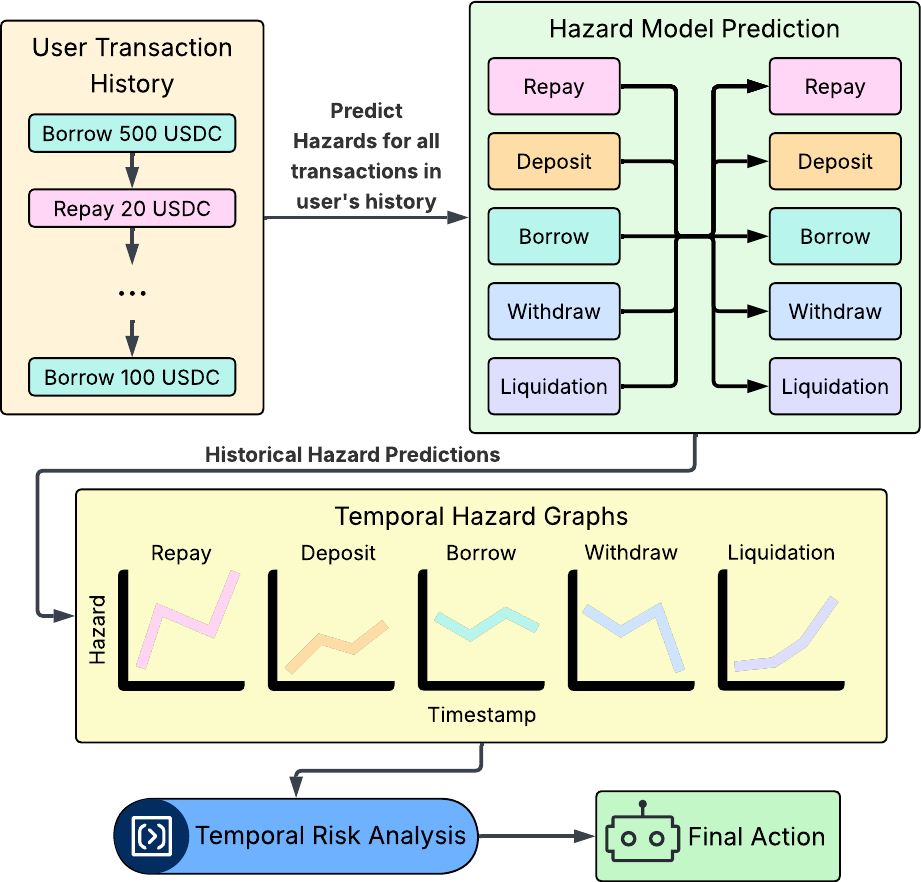}
    \caption{Overview of our agentic liquidation-prevention action determination pipeline. 
    }

    \label{fig:overall}
\end{figure}

In practice, our framework operates as a proactive agent that continuously monitors user portfolios. Unlike static rule-based bots, it leverages counterfactual survival analysis to quantify risk and optimize interventions. The overall architecture is illustrated in Figure \ref{fig:overall}, and the workflow follows a sequential pipeline:

\begin{enumerate}
    \item \textbf{Data Ingestion \& Hazard Prediction:} As shown in the top-left of Figure \ref{fig:overall}, the system ingests the raw user transaction history and feeds it into a hazard model, which predicts the instantaneous risk (hazard) for all competing event types (\texttt{Repay}, \texttt{Deposit}, \texttt{Borrow}, \texttt{Withdraw}, and \texttt{Liquidation}) simultaneously, rather than looking at \texttt{Liquidation} in isolation.
    \item \textbf{Temporal Aggregation:} These raw predictions are aggregated into \emph{Temporal Hazard Graphs} (Figure \ref{fig:overall}, bottom), which visualize how the risk of each event evolves over time. This allows the agent to distinguish between noise and genuine risk trends.
    \item \textbf{Risk Analysis \& Action:} Finally, the \emph{Temporal Risk Analysis} module evaluates these trends to generate a final action, prioritizing interventions only when the \texttt{Liquidation} risk curve dominates the user's organic behavior patterns.
\end{enumerate}

\noindent The core algorithmic contributions enabling this pipeline are: 
\begin{enumerate}
    \item Absolute Risk Quantification using a robust Breslow estimator. 
    \item Temporal Trend Analysis to distinguish signals from noise. 
    \item Counterfactual Action Optimization.
\end{enumerate}

For the predictive core, we utilize XGBoost with the Cox proportional hazards objective (\texttt{survival:cox})~\cite{cox1972regression}. We selected this model for its computational efficiency and proven performance on financial survival tasks (notably achieving second place~\cite{jian2025autofinsurv} in the FinSurvival 2026 Challenge~\cite{finsurvival_challenge,seneviratne2026benchmarking}), which utilized dataset structures similar to ours. This lightweight architecture enables rapid inference across thousands of user profiles, a critical requirement for real-time DeFi monitoring.

\subsection{Absolute Risk Quantification via Return Period}
Standard ``Mean Time to Event" metrics are ill-suited for rare financial events like liquidations, as the integral of the survival function often extrapolates to infinite horizons for safe users, resulting in indistinguishable scores. Additionally, raw outputs from Cox proportional hazards models (log-partial hazards) are relative and baseline-dependent, making it difficult to compare risks across different transaction types (e.g., \textit{\texttt{Deposit}} vs. \texttt{Liquidation}).

To solve this, we introduce a \textbf{Return Period} metric, $T_{ret}$, defined as the inverse of the event probability over a fixed short-term horizon (e.g., $\Delta t = 7$ days), scaled by that horizon. This maps all risks onto a unified, linear temporal scale:
\begin{equation}
    T_{ret}(x) = \frac{\Delta t}{1 - \exp\left(-\hat{H}_0(\Delta t) \cdot e^{\hat{\beta}^T x - S}\right)}
\end{equation}
where $\hat{H}_0(\Delta t)$ is the baseline cumulative hazard at the horizon, $\hat{\beta}^T x$ is the raw log-partial hazard, and $S$ is a stability shift factor (explained below). A high-risk user might have a return period of 2 days, while a safe user has a return period of 10 years, allowing the agent to rank disparate risks directly.

\subsection{Numerically Stable Baseline Estimation}
Standard Breslow estimators often suffer from floating-point overflow when applied to financial data with extreme feature values, leading to ``exploding hazard" errors~\cite{breeden2015instabilities}. We address this with two key modifications in Algorithm \ref{alg:return_period}. In our notation, the training dataset $\mathcal{D}$ consists of tuples $\{(x_i, \delta_i, \tau_i)\}$, where $x_i$ is the feature vector for user $i$, $\delta_i \in \{0,1\}$ is the event indicator (1 if the event occurred, 0 if censored), and $\tau_i$ is the observed time duration until the event or censoring.

\begin{enumerate}
    \item \textbf{Log-Hazard Centering:} We compute a shift factor $S = \text{median}(\text{log-hazards})$ from the training set. By subtracting $S$ before exponentiation, we center the relative risk scores around 1.0, preventing numerical overflow for high-risk users and underflow for safe users.
    \item \textbf{Vectorized Risk Set Calculation:} Instead of iterating through time steps (O($N^2$)), we group data by unique durations and use a \textbf{Reverse Cumulative Sum} to calculate the risk set size $\mathcal{R}(t)$. This ensures numerical stability even when the risk set spans millions of transactions.
\end{enumerate}

\begin{algorithm}[tbh]
\caption{Stable Return Period Estimation}
\label{alg:return_period}
\begin{algorithmic}[1]
\REQUIRE Model $\mathcal{M}$, Training Data $\mathcal{D}=\{(x_i, \delta_i, \tau_i)\}$, Target $x_{new}$, Horizon $\Delta t$
\\ \textbf{Phase 1: Baseline Estimation (Training)}
\STATE Predict log-hazards $\eta_i = \mathcal{M}(x_i)$ for all $i \in \mathcal{D}$
\STATE Compute stability shift: $S \leftarrow \text{median}(\{\eta_i\})$
\STATE Compute shifted Relative Risks: $r_i \leftarrow \exp(\eta_i - S)$
\STATE Group data by unique event times $t_{(1)} < \dots < t_{(K)}$
\STATE Compute Risk Set Sums (Reverse Cumulative):
\\ \quad $\mathcal{R}(t_{(k)}) \leftarrow \sum_{j: \tau_j \ge t_{(k)}} r_j$
\STATE Compute Baseline Hazard Increments:
\\ \quad $\Delta H_0(t_{(k)}) \leftarrow \frac{\sum_{j \in D_{(k)}} \delta_j}{\mathcal{R}(t_{(k)})}$
\STATE Integrate: $H_0(t) \leftarrow \text{cumsum}(\Delta H_0)$
\\ \textbf{Phase 2: Inference}
\STATE Predict user log-hazard: $\eta_{new} = \mathcal{M}(x_{new})$
\STATE Compute user Relative Risk: $r_{new} = \exp(\eta_{new} - S)$
\STATE Look up baseline hazard at horizon: $h^* \leftarrow H_0(\Delta t)$
\STATE Compute Event Probability: $P \leftarrow 1 - \exp(-h^* \cdot r_{new})$
\RETURN $T_{ret} \leftarrow \frac{\Delta t}{\max(P, \epsilon)}$
\end{algorithmic}
\end{algorithm}

\subsection{Volatility-Adjusted Trend Analysis}
\label{sec:trend_analysis}
To distinguish genuine risk accumulation from transient market noise, we implement a \textbf{Dimensionless Trend Score} (Algorithm \ref{alg:trend_slope}). Raw linear slopes are insufficient because risk probabilities fluctuate with varying volatilities across different users. We normalize the linear trend, calculated with Ordinary Least Squares (OLS), by the series' standard deviation ($\sigma_Y$) and time span ($\Delta T$), effectively creating a ``Z-score of change'' ($Z_{slope}$) that represents how many standard deviations the risk has shifted over the window.

We augment this base signal with an \textbf{Acceleration} term ($Z_{accel}$), calculated as the difference between the normalized slopes of the second half ($Z_{recent}$) and first half ($Z_{past}$) of the window. Finally, we apply a penalty based on \textbf{Residual Volatility} ($Z_{vol}$) and modulate the score using a \textbf{Momentum Adjustment} that scales the trend based on the current risk's relative deviation ($Rel$) from the historical mean. The final score is given by:

\begin{equation}
\label{eq:v}
    \mathcal{V} = (Z_{slope} + \omega \cdot Z_{accel} - \lambda \cdot Z_{vol}) \cdot (1 + \gamma \cdot Rel)
\end{equation}

\noindent where $\omega$, $\lambda$, and $\gamma$ represent the weighting coefficients for acceleration, volatility penalty, and momentum adjustment, respectively.
In our experiments, we set these hyperparameters to $\omega=0.8$, $\lambda=0.6$, and $\gamma=0.3$. To confirm these values do not create a fragile decision boundary, a sensitivity analysis yielded a prediction stability score of 0.9353, demonstrating the agent's risk predictions are highly robust to parameter perturbation. The high acceleration weight (0.8) prioritizes proactive responses to fast-moving threats. The volatility penalty (0.6) acts as a noise filter, suppressing mean-reverting transient fluctuations in high-variance assets. Finally, the momentum scalar (0.3) scales intervention urgency based on deviations from the user's historical baseline. This composite metric ensures the agent acts on structural, accelerating insolvency rather than market noise.

\begin{algorithm}[tbh]
\caption{Dimensionless Trend-Risk Scoring}
\label{alg:trend_slope}
\begin{algorithmic}[1]
\REQUIRE Return Periods $Y = \{y_1, \dots, y_n\}$ at times $T = \{t_1, \dots, t_n\}$
\STATE \textbf{Initialize:} $\sigma_Y \leftarrow \text{std}(Y)$, $\Delta T \leftarrow t_n - t_1$
\\ \textbf{Global Trend ($Z_{slope}$):}
\STATE Calculate OLS slope $\beta_{raw}$ on $(T, Y)$
\STATE Normalize: $Z_{slope} \leftarrow \beta_{raw} \cdot \frac{\Delta T}{\sigma_Y}$
\\ \textbf{Acceleration ($Z_{accel}$):}
\STATE Set split length $k \leftarrow \max(2, \lfloor n/2 \rfloor)$
\STATE $Z_{past} \leftarrow \text{Slope}(T_{1:n-k}, Y_{1:n-k}) \cdot \frac{t_{n-k}-t_1}{\sigma_Y}$
\STATE $Z_{recent} \leftarrow \text{Slope}(T_{n-k+1:n}, Y_{n-k+1:n}) \cdot \frac{t_n - t_{n-k+1}}{\sigma_Y}$
\STATE $Z_{accel} \leftarrow Z_{recent} - Z_{past}$
\\ \textbf{Volatility Penalty ($Z_{vol}$):}
\STATE Calculate residuals $R \leftarrow Y - (\beta_{raw}T + \text{intercept})$
\STATE $\sigma_{resid} \leftarrow \text{std}(R)$
\STATE $Z_{vol} \leftarrow \frac{\sigma_{resid}}{\sigma_Y}$
\\ \textbf{Composite Score:}
\STATE $\mathcal{V} \leftarrow Z_{slope} + \omega \cdot Z_{accel} - \lambda \cdot Z_{vol}$
\\ \textbf{Momentum Adjustment:}
\STATE Calculate relative deviation: $Rel \leftarrow \frac{y_n - \text{mean}(Y)}{|\text{mean}(Y)|}$
\STATE $\mathcal{V} \leftarrow \mathcal{V} \cdot (1 + \gamma \cdot Rel)$
\RETURN $\mathcal{V}$
\end{algorithmic}
\end{algorithm}

\subsection{Counterfactual Action Optimization}
The agent employs a \textbf{Counterfactual Optimization Loop} (Algorithm \ref{alg:recommendation}) to determine the \textit{Minimum Viable Intervention} required to prevent liquidation and restore portfolio safety. Unlike rule-based bots that enforce static safety buffers, our agent defines risk \textit{relatively}: a user is considered ``At Risk'' if the predicted \emph{Liquidation Return Period} ($T_{liq}$) is shorter than that of any other event (e.g., $T_{repay}$, $T_{deposit}$, $T_{withdraw}$, $T_{borrow}$), or if the \emph{Liquidation Trend Score} ($\mathcal{V}_{liq}$) is the lowest among all monitored events.

To select the optimal intervention strategy, the agent mimics the user's most likely natural behavior. If the user is at risk, the agent compares the predicted time horizons of organic positive actions: if $T_{repay} < T_{deposit}$, it prioritizes a \texttt{Repay} action; otherwise, it selects \texttt{Deposit}. If the user is not currently at risk, the agent defaults to a \texttt{Deposit} strategy to maximize capital efficiency. In the risk mitigation scenario, the optimization loop iteratively increases the action amount $\alpha$ (starting from a minimal unit) by a factor $m$ and simulates the counterfactual state. The loop terminates when the intervention is sufficient to flip the relative risk metric, ensuring that \texttt{Liquidation} is no longer the most imminent or highest-trend event. We choose to multiply the action amount to avoid excessive iterative loops, as stacked multiplications allow for exponential growth if the intervention continues to fail. We use a factor of $2$ in the experiments.

\begin{algorithm}[tbh]
\caption{Counterfactual Risk Optimization}
\label{alg:recommendation}
\begin{algorithmic}[1]
\REQUIRE Action $\mathcal{A}$
\\ \textbf{Risk Identification}
\STATE Events $\mathcal{E} \leftarrow \{Liq, Repay, Deposit, Withdraw, Borrow\}$
\STATE History $\mathcal{H} \leftarrow \text{getHistory}(\mathcal{A})$
\FOR{each event $e \in \mathcal{E}$}
    \STATE $T_e \leftarrow \text{ReturnPeriod}(e, \mathcal{A})$
    \STATE $\mathbf{T}_{hist} \leftarrow \text{ReturnPeriod}(e, \mathcal{H}_i) \text{ for }\mathcal{H}_i\text{ in }\mathcal{H}$
    \STATE $\mathcal{V}_e \leftarrow \text{TrendScore}(\mathbf{T}_{hist})$
\ENDFOR
\STATE $Risk \leftarrow (T_{liq} == \min_{e \in \mathcal{E}}(T_e)) \lor (\mathcal{V}_{liq} == \min_{e \in \mathcal{E}}(\mathcal{V}_e))$

\textbf{Select Strategy}
\IF{\NOT $Risk$}
    \RETURN \textbf{Recommend} $\{\text{\textit{Deposit}}\}$
\ELSE
    \IF{$T_{repay} < T_{deposit}$}
        \STATE $ActionType \leftarrow \text{\textit{Repay}}$
    \ELSE
        \STATE $ActionType \leftarrow \text{\textit{Deposit}}$
    \ENDIF
\ENDIF

\textbf{Optimize Intervention Amount}
\STATE $\alpha \leftarrow \alpha_{min}$ \COMMENT{Smallest currency unit}
\WHILE{$\alpha \le \text{MaxPossible}(\mathcal{S}, ActionType)$}
    \STATE $\mathcal{S}' \leftarrow \text{Apply}(\mathcal{S}, ActionType, \alpha)$
    
    \FOR{each event $e \in \mathcal{E}$}
        \STATE $T'_e \leftarrow \text{ReturnPeriod}(e, \mathcal{S}')$
        \STATE $\mathcal{V}'_e \leftarrow \text{TrendScore}(\text{Append}(\mathbf{T}_{hist}, T'_e))$
    \ENDFOR
    
    \STATE $Risk \leftarrow (T'_{liq} == \min_{e}(T'_e)) \lor (\mathcal{V}'_{liq} == \min_{e}(\mathcal{V}'_e))$
    
    \IF{\NOT $Risk$}
        \RETURN \textbf{Recommend} $\{ActionType, \alpha\}$
    \ENDIF
    
    \STATE $\alpha \leftarrow \alpha \times m$ 
\ENDWHILE

\RETURN \textbf{Recommend} $\{ActionType, \text{MaxPossible}\}$
\end{algorithmic}
\end{algorithm}

\section{Experimental Setup}
\label{sec:experimental_setup}

\subsection{Dataset and Evaluation Cohort}
\label{sec:dataset}
We utilize a large-scale dataset of over 21.8 million records from the Aave v3 protocol on the Polygon subgraph, covering user transactions and market states. The dataset contains 90 engineered features across three primary categories:
\begin{itemize}
    \item \textbf{User-Centric:} Aggregate statistics such as \textit{userBorrowCount}, \textit{userDepositSumUSD}, and \textit{userActiveDaysMonthly}.
    \item \textbf{Market-Level:} System-wide metrics including \textit{marketLiquidationCount} and total pool liquidity.
    \item \textbf{Transactional \& Temporal:} Features capturing specific transaction attributes (e.g., \textit{amountUSD}) and cyclical time patterns (e.g., \textit{cosTimeOfDay}).
\end{itemize}

Crucially, our dataset builds upon the FinSurvival benchmark~\cite{green2025finsurvival} by explicitly modeling survival as a transition between an \textbf{Index Event} (start of observation) and an \textbf{Outcome Event} (end of observation). This framework defines survival tasks as pairs $(E_{start}, E_{end})$, where the model predicts the time $\Delta t$ until $E_{end}$ occurs given that $E_{start}$ just happened. 

We extend the standard FinSurvival task definition in two key ways:
\begin{enumerate}
    \item \textbf{Liquidation as Index Event:} We include \texttt{Liquidation} as a possible starting event (e.g., \texttt{Liquidation} $\to$ \texttt{Deposit}). This allows us to analyze post-liquidation behavior and recovery, which is critical for understanding user churn.
    \item \textbf{Self-Transitions:} We model transitions where the index and outcome events are identical (e.g., \texttt{Deposit} $\to$ \texttt{Deposit}). This captures recurring user habits and high-frequency interaction patterns that are vital for short-term risk assessment.
\end{enumerate}

\subsection{Aave v3 Simulator Design}
Evaluating proactive liquidation-prevention strategies requires answering counterfactual questions such as: \emph{``would a liquidation have occurred if a different action had been taken earlier?''} that cannot be observed on-chain. For this purpose, we developed a protocol-faithful Aave v3 simulator to serve as the ground-truth environment for validating our agent's action.
The simulator reconstructs user portfolio states by replaying historical transactions. It faithfully implements Aave v3's two-slope interest rate model to accrue interest at every timestep. Key components include:

\subsubsection{Robust Liquidation Inference}
\texttt{Liquidation} is triggered when the HF drops below 1.0. However, on-chain data often lacks the granularity to explain all liquidations due to oracle latency or rapid price changes. We implement a \emph{dynamic margin threshold} that expands the effective liquidation boundary (e.g., $HF < 1.10$) as the time gap between user interactions increases, accounting for unobserved volatility. Additionally, we explicitly model ``dust liquidation'' scenarios, where the absolute value of a user's collateral falls below a protocol-defined minimum (e.g., dust thresholds for specific assets), triggering liquidation even if the HF appears nominally safe. Our detection logic identifies these events by checking if the liquidated debt amount covers nearly 100\% of the user's outstanding balance (debt-to-cover ratio $\approx 1.0$) for very small positions, ensuring these administrative cleanups are correctly distinguished from insolvency events.

\subsubsection{Global Price State Reconstruction}
To ensure accurate collateral valuation, the simulator relies on a globally synchronized price history extracted from the aggregate user dataset. Although simulations are executed on individual user profiles for parallel efficiency, every simulation step at timestamp $t$ queries a unified price oracle that reflects the exact market state of all assets (e.g., WBTC, USDC, WETH) at that historical moment. This prevents ``price drift" between parallel simulations and ensures that liquidation triggers are consistent with the broader market conditions observed on-chain.

\subsubsection{Wallet Balance Inference}
\label{sec:wallet_inference}
Accurately simulating user transactions on Aave v3 requires reasoning not only about protocol-level state (collateral, debt, and HF), but also about whether a user \emph{could have executed} a given transaction given their external wallet balances. While such balances can in principle be obtained from on-chain data, doing so at scale is prohibitively expensive.

To avoid this overhead, we adopt a wallet balance inference approach that estimates users’ initial wallet balances directly from their Aave transaction history. The core idea is to infer the \emph{minimum wallet balance} that must have existed to make the observed sequence of transactions feasible. The algorithm processes transactions in chronological order: deposits and repayments require sufficient balance, while withdrawals and borrows add to it. At each step, we track the maximum negative balance encountered and infer that the user must have started with at least that amount. To account for unobserved inflows (e.g., transfers from exchanges), we apply a multiplicative safety factor (default 1.5x) to the inferred minimums. This strikes a pragmatic balance between realism and scalability, enforcing meaningful constraints without assuming infinite liquidity.

\subsection{Evaluation Methodology: Replay-Based Validation}
To rigorously assess the agent's effectiveness, we employ a replay-based validation strategy that splits user profiles into ``historical'' (used for training/prediction) and ``future'' (used for validation) segments.

\subsubsection{Replay Framework}
For each action generated at timestamp $t_{rec}$, we fork the user's state into two parallel simulations:
\begin{enumerate}
    \item \textbf{Baseline (Control):} The simulator replays the user's actual historical future transactions starting from $t_{rec}$, without intervention.
    \item \textbf{Intervention (Experiment):} The simulator injects the recommended action (e.g., \textit{\texttt{Repay} 500 USDC}) at $t_{rec}$, then replays the same future transaction sequence.
\end{enumerate}
This paired simulation isolates the causal effect of the agent's action on liquidation outcomes.

\subsubsection{Evaluation Metrics}
We measure performance using three key metrics derived from the simulation pairs, specifically tailored to the high-stakes nature of our evaluation cohort:
\begin{itemize}
    \item \textbf{High-Risk Salvage Rate:} The percentage of cases where the user was liquidated in the baseline simulation ($Liquidated_{base}$), but the agent successfully prevented it in the intervention simulation ($\neg Liquidated_{intervention}$). Given that our cohort consists of imminent-risk profiles, this metric effectively measures the agent's ability to ``save the unsavable."
    \item \textbf{Safety (Worsening Rate):} The percentage of cases where the user was \textit{not} liquidated in the baseline, but the agent's intervention \textit{caused} a liquidation ($Liquidated_{intervention} \land \neg Liquidated_{base}$). Ideally, this should be 0\%.
    \item \textbf{Efficiency (Dust Avoidance):} We evaluate the agent's ability to distinguish between economic insolvency and administrative dust liquidations. A successful agent should ignore dust liquidation events where the cost of intervention (gas) exceeds the value at risk.
\end{itemize}


\begin{figure*}[t]
    \centering
    \includegraphics[width=\linewidth]{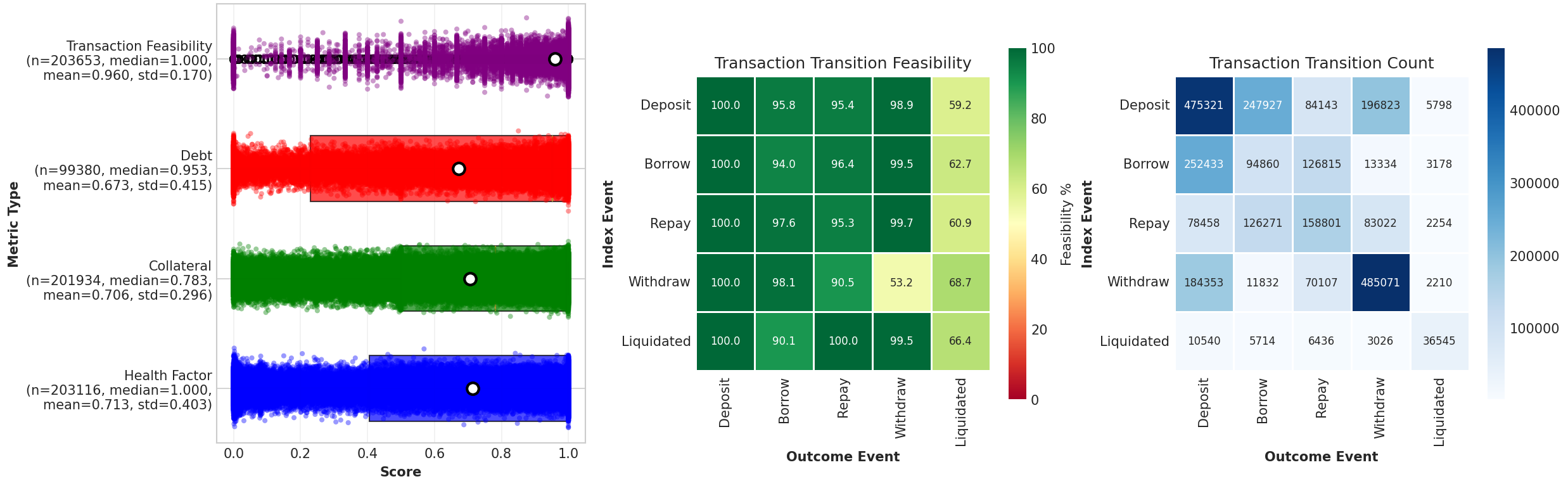}
    \caption{Simulator state fidelity and transaction feasibility.}
    \label{fig:correlation}
    \vspace{-4mm}
\end{figure*}

\begin{figure*}
    \centering
    \includegraphics[width=\linewidth]{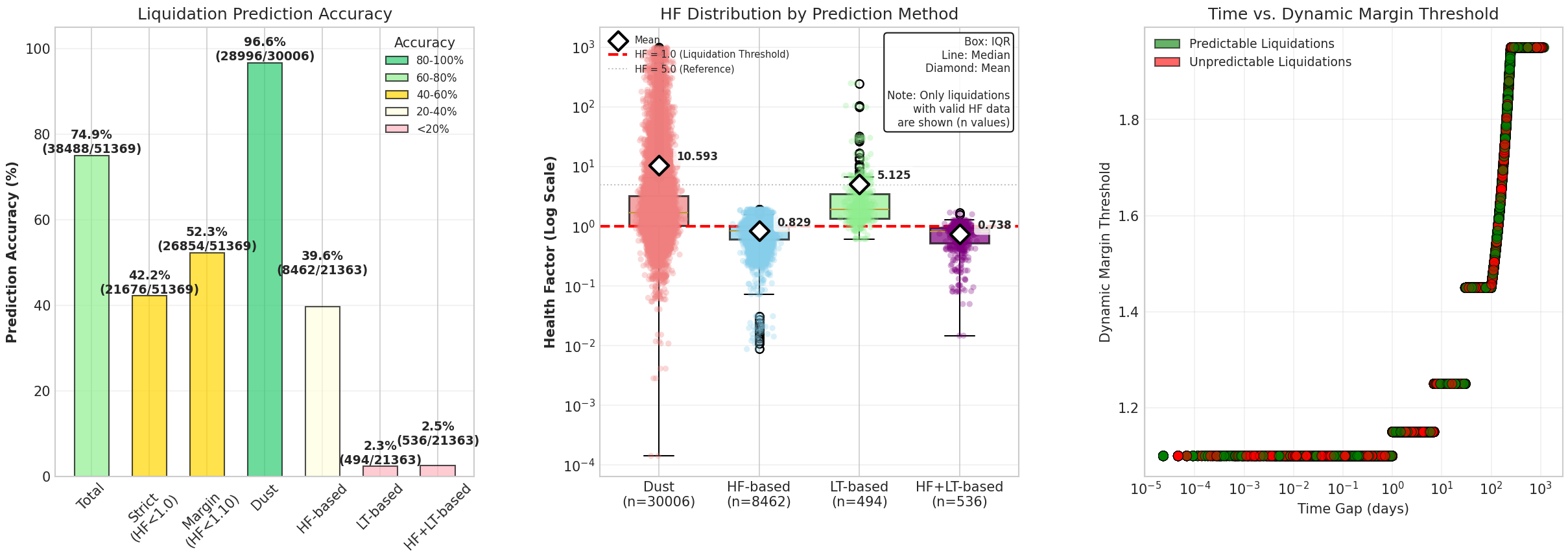}
    \caption{Liquidation prediction behavior under dynamic risk modeling.    
}
    \label{fig:liquidation_prediction_behavior}    
    \vspace{-6mm}
\end{figure*}

\subsection{Aave v3 Simulator Validation Results}
We stress-tested the simulator using a \emph{liquidation-focused split} of the dataset described in \Cref{sec:dataset}, holding out all data following a user's first liquidation. As shown in \Cref{fig:correlation}, the simulator achieves a median correlation of 1.0 with ground-truth HFs. Furthermore, \Cref{fig:liquidation_prediction_behavior} demonstrates that our dynamic margin logic correctly anticipates 74.9\% of liquidation events, compared to only 42.2\% for a strict $HF < 1.0$ rule. This high fidelity ensures that if our agent prevents a liquidation in the simulator, it is highly likely to have prevented it in reality.

\subsection{Agent Evaluation Subset}
To rigorously evaluate our agent designed for liquidation prevention, we constructed a specific \textbf{Evaluation Cohort} from our dataset described in \Cref{sec:dataset}. We selected a time window covering the 40\textsuperscript{th} to 80\textsuperscript{th} percentile of the protocol's history (approx. 2023--2024) to ensure sufficient training history while reserving the final 20\% for future validation. Given the significant computational resources required for high-fidelity agent simulations, evaluating the entire dataset was infeasible. Therefore, we employed a stratified sampling strategy to balance distinct user coverage with computational efficiency:
\begin{enumerate}
    \item For standard event pairs (e.g., \texttt{Deposit} $\to$ \texttt{Withdraw}), we randomly sampled 300 instances per pair to ensure balanced coverage of typical user behaviors.
    \item For critical pairs ending in \texttt{Liquidation} (e.g., \texttt{Borrow} $\to$ \texttt{Liquidation}), we explicitly selected the 300 instances with the shortest \textit{time-to-event} ($\Delta t$). This selection bias intentionally exposes the agent to the most imminent and high-risk scenarios, effectively ``stress-testing'' its ability to detect and mitigate liquidations under extreme time pressure.
\end{enumerate}
This process yielded the 8400 distinct user checkpoints used in our results.

\subsection{Liquidation Detection Engine}
\label{sec:liq}
Validating liquidation events in simulation is non-trivial due to the discrete nature of time-stepping. To ensure robustness, we define a ``Confirmed Liquidation" only when a consensus is reached among six parallel detection algorithms:
\begin{itemize}
    \item \textbf{Event-Driven:} Checks at every price update timestamp.
    \item \textbf{Adaptive Granularity:} Varies check frequency based on Health Factor proximity to 1.0.
    \item \textbf{Binary Search:} Recursively finds liquidation seconds.
    \item \textbf{Hybrid:} Event-driven checks with periodic safeguards.
    \item \textbf{Model-Based:} Checks predicted high-risk windows.
    \item \textbf{Interest Milestones:} Captures slow-bleed insolvencies driven by debt accrual.
\end{itemize}

\vspace{-2mm}
\subsection{Dust Exclusion and Action Feasibility}
\vspace{-1mm}
When evaluating liquidation mitigation, it is crucial to distinguish between \emph{insolvency events} (caused by under-collateralization due to price drops) and \emph{dust liquidations} (administrative cleanups of negligible balances, typically $<\$1.00$). While dust events constitute a significant portion of protocol activity, they do not represent genuine financial risk to the user. To maintain the economic integrity of our evaluation, we filter out dust liquidations.

Additionally, to ensure a fair evaluation of the agent's capabilities, we apply two further exclusions to the test set. First, we remove scenarios where the liquidation occurred so rapidly that no action could physically be generated and confirmed on-chain (e.g., flash crashes within the block time). Second, we exclude anomalous data points where liquidations are recorded despite the user having zero debt (likely protocol artifacts or data ingestion errors), as no financial intervention can logically prevent them.
We assume standard mainnet gas costs for intervention feasibility.

To ensure that the agent's interventions are realistic and executable on-chain, we implement rigorous validation logic. This function enforces two critical constraints:
\begin{enumerate}
    \item \textbf{Wallet Feasibility:} The agent verifies that the user's inferred wallet balance (as derived in Section~\ref{sec:wallet_inference}) is sufficient to cover the recommended repayment amount. If the balance for the target asset is insufficient, the agent attempts to substitute it by sweeping the maximum available balance of the user's largest holding asset, ensuring that actions are not only theoretically optimal but also practically feasible.
    \item \textbf{Atomic Dust Clearance:} If a calculated partial repayment would reduce a user's debt to a value below the protocol's dust threshold (e.g., reducing a \$50 loan to \$0.50), the agent automatically up-sizes the transaction to fully close the position. This prevents the intervention itself from creating a ``dust" state that would be immediately liquidated by protocol cleaners, ensuring that successful rescues are robust and permanent.
\end{enumerate}

\vspace{-2mm}
\section{Results}
\vspace{-1mm}

After filtering, we evaluated our Agentic Framework on a cohort of 4,882 high-risk user profiles, specifically excluding ``dust'' liquidations to focus on genuine solvency risks. 
Using paired replay-based simulations with and without intervention, we measure the agent's ability to prevent liquidations proactively during the test window.

\subsection{Safety and High-Risk Salvage Rate}

The primary directive of our agent is ``do no harm'' with its generated action. Our simulation confirms that the agent prioritized safety above all else, achieving a \textbf{zero Worsening Rate}. In no instance did the agent's action trigger a liquidation that would not have otherwise occurred. This validates the robustness of the Counterfactual Optimization loop (\Cref{alg:recommendation}), which acts as a safety verify gate before any transaction is proposed.

Beyond safety, the agent achieved a \textbf{Liquidation Reduction Rate of 86.83\%} (1,470 successful rescues out of 1,693 baseline insolvencies). While a double-digit percentage might appear modest in isolation, it is a significant achievement given the adversarial nature of our evaluation cohort. Recall that this cohort was explicitly sampled to include the ``shortest time-to-event'' cases, i.e., users who were essentially ``doomed'' with imminent liquidations, typically occurring within days or hours. In this context, rescuing nearly 87\% of these terminal positions indicates that the agent effectively acts as a high-precision last line of defense where standard buffers had already failed.
From an economic perspective, each prevented liquidation avoids not only a direct penalty (typically 5--10\% of collateral value) but also downstream losses due to forced deleveraging during adverse market conditions, making even modest salvage rates economically significant at scale.

\subsection{Differentiation of Dust vs. Insolvency}

A key finding is the agent's ability to differentiate between administrative noise and financial signal.
\begin{itemize}
    \item \textbf{Insolvency Events (HF-Based):} The agent successfully prevented 1,470 HF-based liquidations. These represent scenarios where market volatility degraded collateral value, and the agent's intervention (\texttt{Repay} or \texttt{Deposit}) successfully restored the HF above the safety threshold.
    \item \textbf{Dust Events:} By design, the agent deprioritized dust liquidations. Our analysis confirms that attempting to rescue dust positions (often $<\$1.00$ in value) is economically inefficient due to gas costs. The agent's implicit filtering of these events optimizes capital allocation towards economically meaningful interventions.
\end{itemize}

\subsection{Action Efficacy and Predictive Accuracy}

Evaluating the performance of liquidation interventions requires understanding both the success of the executed actions and the reliability of the underlying risk signals. Across our expanded cohort of 4,882 profiles, the XGBoost-Cox predictive core achieved an overall accuracy of 69.11\%. This performance confirms the model effectively captures the nuances of varying market regimes. By reliably distinguishing genuine insolvency risks from safe states, the agent ensures efficient capital deployment, averting liquidations while avoiding costly over-reactions to transient market noise.



\subsection{Simulation Robustness}

To ensure these results were not artifacts of simulation noise, we employed the multi-strategy liquidation detection engine detailed in Section \ref{sec:liq}. The results showed \textbf{100\% Consensus Agreement} across all six strategies for both the baseline and intervention simulations. This unanimous agreement confirms that the liquidations prevented by the agent were unambiguous, hard-coded protocol events, and the rescue states were robust across all temporal granularities.

\section{Conclusion}

In this work, we presented an agentic AI framework that reframes DeFi portfolio management as a proactive, time-to-event prediction problem. By replacing static HF rules with a ``Return Period" metric normalized via Cox proportional hazards models, our agent anticipates liquidations before they become imminent. We validated this approach using a novel, protocol-faithful Aave v3 simulator that reconstructs global pricing states and infers wallet feasibility. Empirically, the agent prioritized safety, achieving a zero Worsening Rate while rescuing nearly 87\% of critically at-risk positions that standard buffers failed to save. Furthermore, the system successfully distinguished between actionable insolvency risk and administrative ``dust," optimizing capital efficiency.

More broadly, this work illustrates how predictive blockchain analytics can be elevated into economically grounded, autonomous decision systems. By tightly coupling survival modeling, agentic reasoning, and protocol-faithful simulation, we move toward a new class of financial agents that operate safely under extreme volatility, an essential capability for the next generation of DeFi infrastructure.

\bibliographystyle{IEEEtran}
\bibliography{references}

\end{document}